# A novel algorithm for segmentation of leukocytes in peripheral blood

Haichao Cao, Hong Liu* and Enmin Song
Huazhong University of Science and Technology, School of Computer Science & Technology
E-mails: hl.cbib@gmail.com; dr_caohaichao@163.com; esong@hust.edu.cn

**Abstract**

In the detection of anemia, leukemia and other blood diseases, the number and type of leukocytes are essential evaluation parameters. However, the conventional leukocyte counting method is not only quite time-consuming but also error-prone. Consequently, many automation methods are introduced for the diagnosis of medical images. It remains difficult to accurately extract related features and count the number of cells under the variable conditions such as background, staining method, staining degree, light conditions and so on. Therefore, in order to adapt to various complex situations, we consider RGB color space, HSI color space, and the linear combination of G, H and S components, and propose a fast and accurate algorithm for the segmentation of peripheral blood leukocytes in this paper.

First, the nucleus of leukocyte was separated by using the stepwise averaging method. Then based on the interval-valued fuzzy sets, the cytoplasm of leukocyte was segmented by minimizing the fuzzy divergence. Next, post-processing was carried out by using the concave-convex iterative repair algorithm and the decision mechanism of candidate mask sets. Experimental results show that the proposed method outperforms the existing non-fuzzy sets methods. Among the methods based on fuzzy sets, the interval-valued fuzzy sets perform slightly better than interval-valued intuitionistic fuzzy sets and intuitionistic fuzzy sets.

# 1. Introduction

Medical image processing occupies a leading position in the hematology, a branch of medical science that focuses on the study of blood and hematopoietic tissue diseases. In pathological studies, the number of erythrocytes (red blood cells), leukocytes (white blood cells), platelets and other blood cells is very important for the detection of diseases such as anemia, leukemia, cancer and other infectious diseases. Among those blood cell parameters, the number of leukocyte plays an essential role in the body's immune system [1, 2]. Blood leukocytes are mainly divided into five categories, namely eosinophils, basophils, neutrophils, lymphocytes and monocytes. The five types of leukocytes can be identified by the size of cell, the type of nucleus lobes, the ratio of nucleus to cytoplasm, cytoplasmic granules and the staining properties of granules, etc.[3].

Particularly the numbers and types of leukocytes are essential for the detection of blood type disease. Traditional medical experts count the cells directly, though it is not only quite time-consuming but also frequent to miss and repeat. Nowadays, automated techniques have been introduced for the diagnosis of medical images [4]. However, under different backgrounds, methods of staining, and degrees of dyeing, light conditions and so on, it remains difficult to accurately extract related features and count the number of cells [5, 6]. Therefore, in order to adapt to various complex situations, a method based on stepwise averaging method, interval-valued fuzzy sets and concave-convex iterative repair algorithm for segmentation of leukocytes is proposed in images of Wright's stained [7, 8] blood smears preparations.

In recent years, many researchers have proposed different segmentation solutions for blood cells. We divide the leukocyte segmentation method into two categories, i.e., one is based on non-fuzzy set methods and the other is based on fuzzy set methods.

The non-fuzzy set approaches for leukocytes segmenting are based on threshold [9-11], morphological operations [12], region growing [13], neural network[14], and clustering [15-17]. These methods, however, may lead to unsatisfactory segmentation results due to the existence of the red blood cells, and its performance is not good for overlapping cells or cells with boundaries that are not smooth enough (rough edge). Some methods have been developed based on deformable models, such as watershed algorithm [18, 19], level set [20] and parametric active contour [3, 21]. However, these methods require good contour initialization to obtain good segmentation results, and the processing time is long, and cannot meet the requirement of engineering application. In addition, several methods that can be categorized into color-based methods have been proposed in the literature to exploit the advantages of the color variation between different cell components in segmentation [22-27]. However, these methods in [23-25] are unable to segment the cytoplasm, and the nuclei segmentation results still need to be improved. There are also some recent approaches proposed, for example, in [28], the authors uses multiple windows obtained by scoring multiscale cues to locate the leukocytes, and then the GrabCut algorithm based on dilation is iteratively run to segment the leukocytes.

In the fuzzy set approaches, the fuzzy divergence is used to obtain the optimal threshold. In [29], the authors proposed a Yager's method based on fuzzy sets, which respectively processed three channel images of R, G and B to get three segmented images. Finally, the three segmentation images are merged to obtain the final segmentation results. However, it cannot be applied to the larger connected cells. In [30], the authors proposed an automatic segmentation by intuitionistic fuzzy divergence based thresholding, that can still ensure better robustness in the case of noise interference. However, it is not good for the low staining and weak edge information, and the scale of test sample set is small, which needs to be further tested. In [31, 32], the authors introduced a general version of the classical fuzzy sets [33] known as intuitionistic fuzzy sets. Extending the concept further, the authors used Atanassov's intuitionistic fuzzy sets and interval II fuzzy sets theory to carry out the segmentation of blood leukocyte image [34]. Compared with other algorithms, it has a better performance on the segmentation of an image containing more than one cell. In [35], the authors developed a new leukocyte segmentation methodology based on intuitionistic fuzzy divergence to achieve robust segmentation performance automatically without training set. However, it cannot adapt to changes in background and brightness.

The leukocyte segmentation technique proposed in this paper has the following three innovations:

(1) We propose a novel thresholding method and color transformation methods.

(2) The G, H and S three single-channel images are considered, which improves the adaptability of the algorithm to various backgrounds, and has better segmentation performance on the samples with lighter staining.

(3) In order to deal with the situation of connected cells, an improved concave-convex iterative repair algorithm is proposed.

## 2. Methods

The number and type of peripheral leukocytes are essential for the diagnosis of blood diseases. Because doctors cannot focus their attention for a long time, manual counting and classification are prone to omissions and repetitions. Therefore, there is an urgent need for an automated and accurate leukocyte analysis technique that generally involves three steps, namely cell segmentation, feature extraction, and cell classification. We focus on the first step: cell segmentation, since the accuracy of subsequent steps mainly depend on the cell segmentation in automated leukocyte analysis.

In this paper, the leukocyte segmentation algorithm is divided into three modules: the segmentation of nucleus, sub-image cropping, and the segmentation of cytoplasm. The flow chart is shown in Fig. 1.

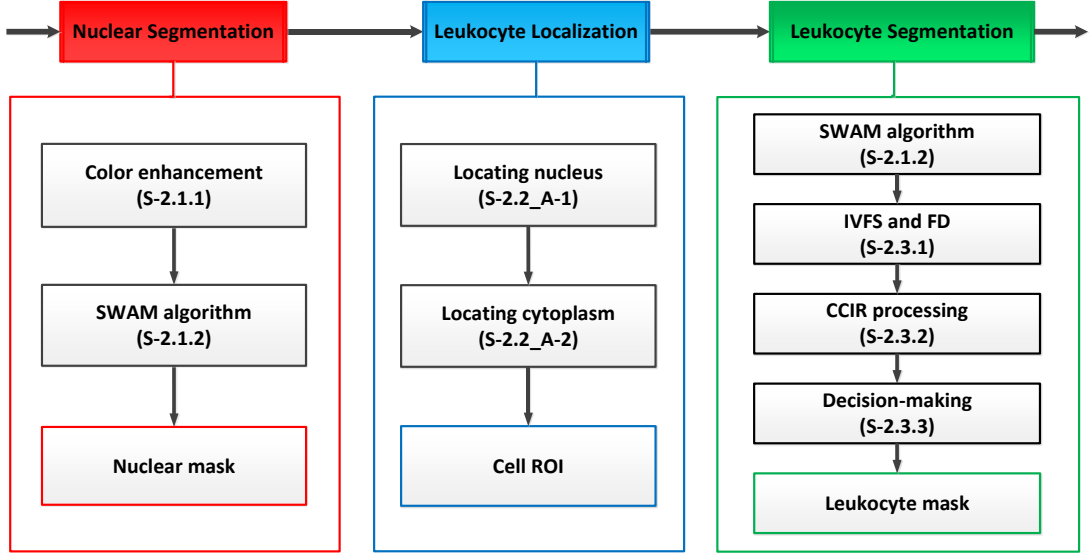

Fig. 1. The flowchart of the proposed algorithm. Three modules are shown within the red, green and blue boxes, respectively. Note that, "S-c" denotes section c; "A-n" denotes algorithm n; "SWAM (stepwise averaging method)" denotes proposed threshold-based segmentation algorithm. For details, see section 2.1.2; "CCIR" denotes a concave-convex iterative repair algorithm; "IVFS" denotes interval-valued fuzzy sets; "FD" denotes fuzzy divergence.

## 2.1 Nuclear segmentation

The segmentation process of leukocyte nucleus is firstly, constructs the transformed image, HSG; secondly, the SWAM segmentation method was used to calculate the thresholds for segmentation of background, erythrocytes, cytoplasm and nucleus respectively, namely $T_b$, $T_r$, $T_c$ and $T_n$. Finally, nucleus can be segmented by using the average of $T_c$ and $T_n$ as the threshold value. Note: the cytoplasm and nucleus are referring to the cytoplasm and nucleus of leukocytes.

The construction of the transformed image HSG and the SWAM segmentation method are described in section 2.1.1 and 2.1.2, respectively.

## 2.1.1 Transformed image

In general, after Wright's staining of the blood smear, the color of nucleus is dark purple, which is darker than that of cytoplasm and mature red blood cells. The blood smear image is usually stored in the RGB (Red-Green-Blue) format. In order to enhance the contrast between the nucleus of leukocytes and other cell tissues, it is necessary to convert from RGB to Hue-Saturation-Intensity (HSI). Our experiments show that the grey value of leukocyte nucleus is lower than non-nucleus pixels in the green channel, and the grey value distribution of leukocyte nucleus is the highest compared with non-nucleus pixels in the Hue (H) component. Similarly, it is also true for the Saturation (S) component. According to this observation, we can derive a color transformation for the leukocyte as shown in Eq. (1).

In Eq. (1), $I_H(x, y)$, $I_S(x, y)$ and $I_G(x, y)$ represent the grey values of a pixel in the coordinate (x, y) in the H channel, S channel and G channel, respectively. Parameters $w_1$, $w_2$, $w_3$ are real values in [0, 1] that can be adjusted to experimental results. In this study, the value of $w_1$, $w_2$, $w_3$ is 0.4, 0.6, and 1.0.

$$I_{HSG}(x, y) = \begin{cases} \dfrac{\omega_1 \times I_H(x, y) + \omega_2 \times I_S(x, y)}{\omega_3 \times I_G(x, y)}, & I_G(x, y) > 0 \\ 255, & I_G(x, y) = 0 \end{cases} \quad (1)$$

The image obtained by Eq. (1) is called the Hue, Saturation, and Green color enhancement image (HSG). In contrast to the image in each of H, S, and G channels, the color enhancement image significantly highlights the region of leukocyte nucleus [36].

## 2.1.2 Stepwise averaging method

The stepwise averaging method (SWAM) is a threshold-based segmentation algorithm. The algorithm works the best if the proportion of non-target areas is larger than the target area. The algorithm takes the average grey value of the image as a threshold. If the grey value of the pixel is smaller than the threshold, it will be unchanged. Otherwise, the grey value of the pixel is set to zero. We can obtain an image after this processing; Secondly, process the image obtained in the previous step in the same way until the mean of the image does not change. Theoretically, the final mean value obtained is the grey value of the background in the bone marrow microscopic image. The background can be removed from the image by using the final mean value. Using this method, we can obtain the grey value of mature red blood cells, the cytoplasm and the nuclei of leukocyte in bone marrow microscopic images. The threshold of segmenting the leukocyte nucleus can be obtained, and the mask image of leukocyte nucleus can be obtained according to this threshold.

In the practical application, SWAM algorithm flow is as follows.

| | |
|---|---|
| **SWAM algorithm** | |
| **Input:** | An enhanced leukocyte image using the color transformation, initial parameter $T_i = 0$ and a number of iterations IT=1. |
| **Output:** | A mask image. |
| **S1:** | Calculate the average grey value of the pixels with the grey value greater than $T_i$ in the HSG image, denoted as $T_j$. |
| **S2:** | Calculate the average grey value of the pixels with the grey value which is less than $T_j$ but greater than $T_i$ in the HSG image, denoted as $T_k$. |
| **S3:** | Set $T_i = T_k$ and IT = IT + 1. If IT = 4, stop. Otherwise, repeat Steps 1 and 2. |
| **S4:** | The four average grey values will be obtained for the background, the mature red blood cells, the leukocyte cytoplasm and nuclei. Then, use the average grey value of leukocyte cytoplasm and nuclei for a rough initial segmentation to obtain a mask image. If a pixel value is less than the average value, it is set 0. Otherwise, it will be set to 255. The mask image obtained will contain the leukocyte nuclei. |

Experimental results of algorithm SWAM applied to HSG image is shown in Fig.2. In this Fig.2, (a) shows the original leukocyte image; (b) the corresponding HSG image; (c) the output of algorithm SWAM, a leukocyte nucleus image that may contain impurities; (d) the output of algorithm SWAM is post-processed to obtain an impurity-free leukocyte nucleus image.

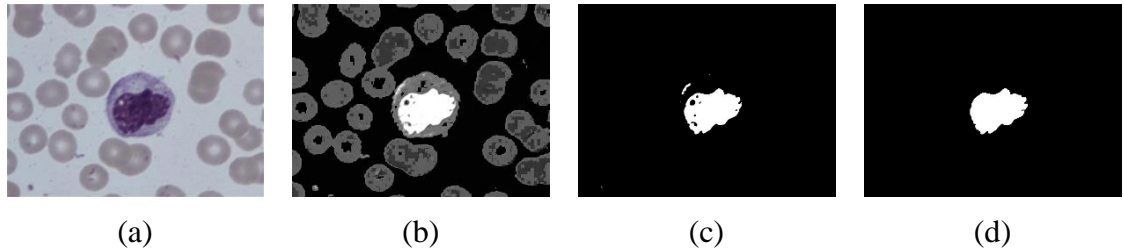

(a)          (b)          (c)          (d)

Fig. 2. (a) the original leukocyte image; (b) the corresponding HSG image; (c) the output of algorithm SWAM, a leukocyte nucleus image that may contain impurities; (d) the output of algorithm SWAM is post-processed to obtain an impurity-free leukocyte nucleus image.

## 2.2 Leukocyte localization

Based on the obtained nuclear mask image, we can identify the region of interest (ROI) of the leukocyte based on the morphological characteristics of the leukocyte. The ROI is a rectangular area, which contains the leukocyte nuclei. We propose different algorithms for locating the leukocyte nuclei and cytoplasm.

The specific algorithm for identifying the ROI of the leukocyte nuclei is shown in Algorithm 1.

**Algorithm 1 (locating the leukocyte nucleus)**

**Input:** The mask image of leukocyte nucleus obtained from SWAM Algorithm.
**Output:** The ROI of the leukocyte nuclei.

**S1:** Obtain the outermost contour of the leukocyte nuclei from the mask image of leukocyte nucleus [37].
**S2:** Calculate the circumscribed rectangle of the contours and its center.
**S3:** Determine whether the two neighboring rectangles need to be combined. If two rectangles are overlapped and two centers are falling inside the overlapped area, the two nuclei will be merged into one.
**S4:** The ROI of the leukocyte nuclei is the circumscribed rectangle of the contours determined by Step 3.

The specific algorithm for identifying the ROI of the leukocyte cytoplasm is shown in Algorithm 2.

**Algorithm 2 (locating the leukocyte cytoplasm)**

**Input:** The mask image of leukocyte nucleus obtained from SWAM Algorithm.
**Output:** The ROI of the leukocyte cytoplasm.

**S1:** Calculate the area, S, and the perimeter, L, of the nucleus.
**S2:** Calculate the circularity of the nucleus $CirR = 4\pi S / L^2$.
**S3:** Calculate the reference radius of the nucleus $R = \sqrt{(S/\pi)} = L/(2\pi)$.
**S4:** Calculate the equivalent radius $R_e$ of the circular region of the cell by

$$R_e = \begin{cases} 2.6 \times R, & CirR < T_1 \\ 2.3 \times R, & T_1 < CirR < T_2 \\ 1.6 \times R, & T_2 < CirR \end{cases} \quad (2)$$

Where the threshold parameters $T_1$ and $T_2$ are given. According to our experiments, it is suggested to use $T_1=0.46$, $T_2=0.85$.
**S5:** The ROI of the leukocyte cytoplasm is the circumscribed rectangle of the circular region determined by $R_e$.

Combining the ROIs of the nucleus and cytoplasm obtained in the two algorithms above by taking the top-left corner and bottom-right corner of two ROIs and re-drawing a new ROI, which will include two smaller ROIs, and we can locate the leukocyte accurately.

Experimental results of leukocyte localization are shown in Fig.3. In this Fig.3, (a) shows the original leukocyte image; (b) the corresponding nucleus image; (c) the output of Algorithm 1; (d) the output of Algorithm 2; (e) the output of combining the ROIs of the nucleus and cytoplasm.

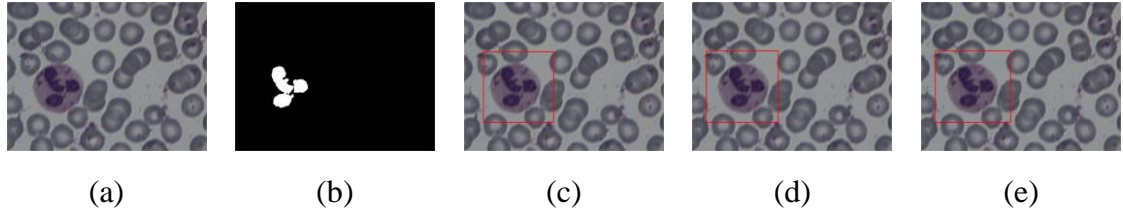

(a)          (b)          (c)          (d)          (e)

Fig. 3. (a) the original leukocyte image; (b) the corresponding nucleus image; (c) the output of Algorithm 1; (d) the output of Algorithm 2; (e) the output of combining the ROIs of the nucleus and cytoplasm.

## 2.3 Leukocyte segmentation

Because the background environment of sample image is very complicated. At least one channel of G, H, and S images in different environments has good properties, which contains a more complete target information. In order to better adapt to various environments, we consider G, H, S three single-channel images. Moreover, taking the candidate image with the highest overall score as the result, the algorithm steps of segmenting the leukocyte are as follows.

(1) According to the SWAM segmentation method, the approximate range of the segmenting thresholds of background, erythrocytes, cytoplasm and nucleus in G, H, S images are determined, recorded as $T_b$, $T_r$, $T_c$ and $T_n$ respectively;

(2) Under the condition that the leukocytes are not connected with red blood cells or other leukocyte, the threshold $T_u = \alpha \times T_b + \beta \times T_r$ can be used to remove the background, where $\alpha$, $\beta$ satisfy $\alpha + \beta = 1$, $\alpha > \beta$;

(3) The segmentation threshold $T_s$ of the cytoplasm of each channel image were determined by the interval-valued fuzzy sets and fuzzy divergence;

(4) The image $I_k, k \in [1, 2]$ is obtained by thresholding the corresponding image with the threshold $T_u, T_s$, and the maximum connected region $M_k$ of the image $I_k$ is selected;

(5) $CCIR$ processing was performed on the image $M_k$ to remove the region that doesn't belong to the leukocytes;

(6) The circularity of $M_k$ is calculated, and the image with larger circularity is pressed into the candidate image set;

(7) After processing G, H and S three single-channel images, making decision for the candidate image set.

Among them, interval-valued fuzzy sets and fuzzy divergence, CCIR processing and decision of candidate image are described in section 2.3.1, 2.3.2 and 2.3.3 respectively.

## 2.3.1 Interval-valued fuzzy sets and fuzzy divergence

In the classical set, each element in universe X have a clear boundary, it either belongs entirely to a set A or not to set A at all, so that the classical set is also called a "Crisp Sets". This phenomenon can be represented by a characteristic function $\phi_A(x)$, which is defined as follows [38].

$$\phi_A(x) = \begin{cases} 1 & x \in A \\ 0 & x \notin A \end{cases} \tag{3}$$

In 1965, Zadeh proposed the concept of fuzzy sets, which extended the characteristic function of the classical set. The range of the feature function is extended from the binary set [34] to the unit interval [0, 1]. This extension function is called the membership function, which depicts the tendency of things in the intermediate transition state to both sides. In this model, every sample point x in the universe X has a degree of membership. However, interval-valued fuzzy set (IVFS) are the fuzzy sets for which the degree of membership $\mu(x)$ does not have a single value for every element but an interval value. Hence, interval-valued fuzzy sets can be obtained by blurring fuzzy sets. That one can define fuzzy sets and assign upper and lower membership degrees to each element known as interval-valued fuzzy sets. An interval-valued fuzzy set $A$ may be written as.

$$A = \{(x, \mu_A^{lower}(x), \mu_A^{upper}(x)) \mid x \in X\} \tag{4}$$

Where $\mu_A^{lower}(x) \in [0,1]$ denotes that the lower probability value of the sample x belongs to $A$, $\mu_A^{upper}(x) \in [0,1]$ denotes that the upper probability value of the sample x belongs to $A$. In addition, $\forall x \in X$ holds for $0 \leq \mu_A^{upper} \leq 1$.

The fuzzy divergence (FD) between two fuzzy sets measures the extent to which the two sets differ from each other. Many divergence formulae have been proposed in the literature [39-41]. One simple divergence formula $D_{AB}(\mu_A, \mu_B)$ between two sets $A$ and $B$ in the universe $X$ is the Hamming distance [16]. It is defined as follows.

$$D_{AB} = \sum_{x \in X} (|\mu_A(x) - \mu_B(x)|) \tag{5}$$

The principle of its application in image segmentation is to find N-1 optimal grey level thresholds $t_k^*, k \in (1, N-1)$, where N is the number of classes in the image. To minimum the fuzzy divergence between the thresholding image with the threshold value $t_k^*$ and the ideal thresholding image. The ideal thresholding image is that each pixel belongs strictly to an exact category. Therefore, the membership degree of each pixel of the ideal thresholding image is one.

The detailed implementation steps for solving the optimal segmentation threshold group $t_k^*$ are as follows.

(1) To found the N-1 optimal thresholds, N-1 layer loops are needed, where $t_k^*$ varying from zero to 255. Then, the value range of $t_k^*$ can be shorten by the SWAM segmentation method, which speed up the running speed of the program.

(2) The grey level images A may be divided into N regions with the determined N-1 thresholds $t_k^*$, denoted as $r_c$, $c \in (1, N)$, $t_0 = 0$, $t_N = 255$.

$$t_{c-1} \leq f(r_c) \leq t_c \tag{6}$$

(3) The average value of the N regions is calculated respectively. The average grey value of the region $r_c$ is defined as follows.

$$avg_c = \frac{\sum_{f=lb(r_c)}^{f=ub(r_c)} f \times count(f)}{\sum_{f=lb(r_c)}^{f=ub(r_c)} count(f)} \tag{7}$$

Where $lb(r_c)$ and $ub(r_c)$ denote respectively the lower and upper bounds of the grey value of region $r_c$, count (f) represents the number of pixels in the image where the grey value is f.

(4) Through the deformation of the Cauchy distribution to determine the upper and lower limits of the membership of each pixel (i, j), the formula is as follows.

$$\mu(i,j) = \frac{1}{[1 + \text{const}(f(i,j) - avg_c)]} \tag{8}$$

The lower limit of the membership of the pixel (i, j) is $\mu^{lower}(i,j) = \mu(i,j)^{1/\delta}$, and the upper limit is $\mu^{upper}(i,j) = \mu(i,j)^{\delta}$. The value of const is $1/(f_{max} - f_{min})$.

(5) Using the basic T - Norms relationship, a new membership degree $\mu^{new}(i,j)$ is generated, which is calculated as follows.

$$\mu^{new}(i,j) = \mu^{lower}(i,j) + \mu^{upper}(i,j) - \mu^{lower}(i,j) \times \mu^{upper}(i,j) \tag{9}$$

(6) Based on the exponential entropy, if the image B is an ideal thresholding image, the fuzzy divergence between images A and B can be defined as follows.

$$D_{IF} = \sum_{i=0}^{M-1}\sum_{j=0}^{N-1}[2 - \{1 - \mu_A(a_{ij}) + \mu_B(b_{ij})\}e^{\mu_A(a_{ij}) - \mu_B(b_{ij})} \\ - \{1 - \mu_B(b_{ij}) + \mu_A(a_{ij})\}e^{\mu_B(b_{ij}) - \mu_A(a_{ij})}] \tag{10}$$

Since image B is an ideal thresholding image, then $\mu_B(b_{ij}) = 1$, so the equation (10) can be simplified as formula (11).

$$D_{IF} = \sum_{i=0}^{M-1}\sum_{j=0}^{N-1}[2 - \{2 - \mu_A(a_{ij})\}e^{\mu_A(a_{ij}) - 1} - \{\mu_A(a_{ij})\}e^{1 - \mu_A(a_{ij})}] \tag{11}$$

(7) Finding a set of optimal thresholds $t_k, k \in [1, N-1]$ for which the divergence $D_{IF}$ is minimized.

(8)Determining the final segmentation threshold $T^*$. For the G channel image, the number of classes $N=4$ and the optimal threshold $T^* = 0.8T_2 + 0.2T_3$. For the H channel image, the number of classes $N=3$ and the optimal threshold $T^* = 0.8T_1 + 0.2T_2$. For the S channel image, the number of classes $N=2$, and the optimal threshold $T^* = T_1$.

## 2.3.2 Concave-convex iterative repair algorithm

The concave-convex iterative repair algorithm is referred to as CCIR algorithm, which is based on finding the pole, to determine whether terminate iteration according to whether the centroid change and the number of poles is less than a given threshold. The solution process of the poles is as follows.

Assuming that the number of pixels on the contour of the leukocyte is L, the centroid of the nucleus is O, the starting point of the contour is S and the cell contour is expressed in the form of polar coordinates $(\rho_i, \theta_i), i \in [1, L]$, where $\rho$ is the distance from the centroid O to the boundary point, $\theta$ represents the angle between the boundary point and the starting point. For a closed contour, the range of $\theta$ is $[0^0, 360^0]$. Based on the two assumption that "the shape of the cell is a circle-like" and "a pair of depressions appear in the approximate polygon of the image when two cells are adhering", $\theta$ should be a monotonically increasing function if the cell is not adhering, otherwise $\theta$ is not monotonic. So when leukocytes are connected with the surrounding cells, the connected region can be determined by detecting the poles.

In the process of finding the pole, the choice of starting point has a certain impact on the determination of the pole. In order to make generally appear as an increasing trend, the steps of selecting the contour starting point are shown in Algorithm 3.

| **Algorithm 3 (determining the starting point of the contour)** |
|---|
| **Input:**      The number L of pixels in the cell contour, the center O of the nucleus, and the move step Sp = 0. |
| **Output:**      The starting point of the contour. |
| **S1:**      Select the contour point pc with the smallest distance from the center O as the initial contour starting point. |
| **S2:**      Determine whether the connection between the contour point pc and the center O has two intersections with the contour. If so, the pc can be the starting point of the contour and the program is terminated. Otherwise, Sp = Sp + 1, the program continues to perform step three. |
| **S3:**      Update the contour point pc = $2^{Sp}$ % L, where the symbol '%' indicates |

|  | modulo operators. |
|---|---|
| **S4:** | Jump to step two. |

After determining the poles, the number of iterations is determined by the pair number of poles. In each iteration process, whether there is a need for repair the two poles is determined according to the number of poles, cell circularity, cell area and other factors before and after the repair. If the pairs of pole need to be repaired, the number of poles should be reduced after repairing, the circularity of the cells should be increased by 5% than before, the area of cells should also be reduced compared to before.

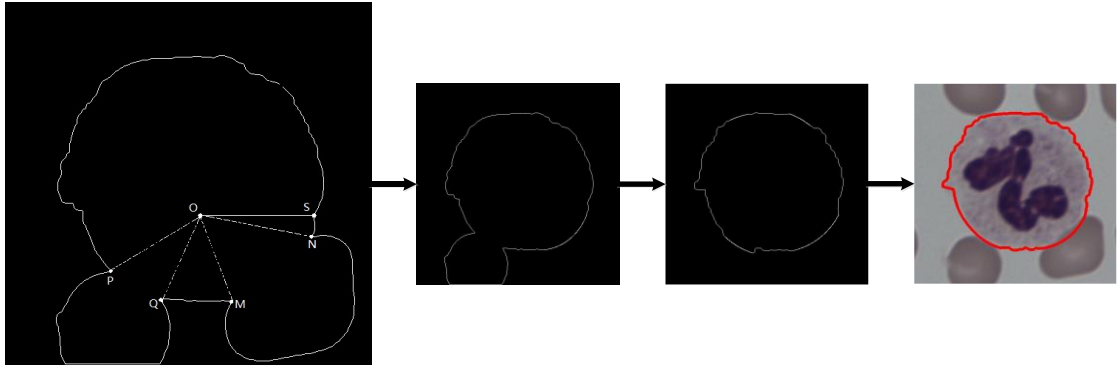

Fig. 4. CCIR iterative process. The first sub-image shows the edge contour of the connected cells. The second sub-image shows the results after repairing the M and the N pole, the third sub-image shows the results after two iterations are completed, and the fourth sub-image shows the final Segment results. Note that, the pole "M, N" in the Fig.5.

In this paper, we use the linear interpolation method to fit the edge curve between two poles that need to be repaired. The concrete process is as follows.

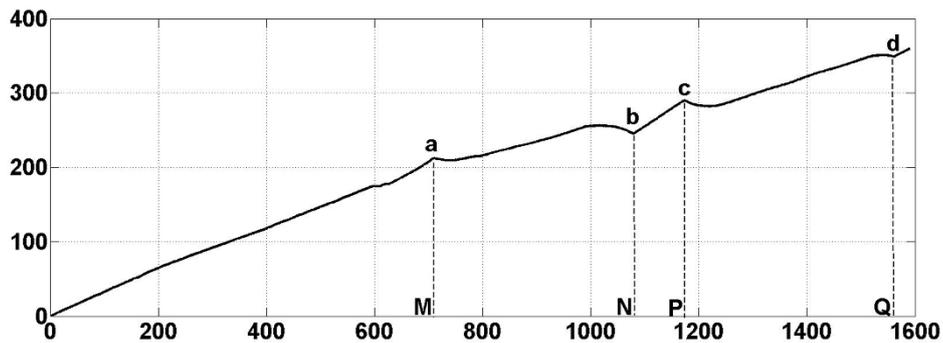

Fig. 5. The relationship between the phase angle and the pixel location

Fig.4 shows the CCIR iterative process. Fig. 5 shows the mapping relationship between the corresponding phase angle and the pixel location. The abscissa represents the ID of the cell contour pixel, and the ordinate represents the phase angle corresponding to each pixel, where the abscissa $M, N, P, Q$ of points $a, b, c, d$ is the ID value of the poles of cell edge contour. To repair M and N poles as an example, suppose $\rho_M < \rho_N$, there are w pixels between the pole M and the pole

N, updating the distance of the w pixels to the centroid O with the idea of linear interpolation $\rho_n, n \in [0, w-1]$, then $\rho_n$ meets the formula (12) [42].

$$\rho_n = \rho_{n-1} + \Delta = \rho_0 + n\Delta \tag{12}$$

In the above formula, $\Delta = (\rho_{w-1} - \rho_0)/(w-1)$, where $\rho_0 \leftarrow \rho_M$, $\rho_{w-1} \leftarrow \rho_N$.

### 2.3.3 Decision of candidate image

In order to improve the accuracy of decision-making as far as possible, we propose four decision parameters: CirRato, BAdh, Sgmv, CirSim, its specific meaning is as follows.

(1) The circularity of leukocyte is referred to as CirRato. The candidate mask image MaskImg is considered as a circle and the parameter CirRato is defined by the relationship between the area and the circumference of the circle. Since the cells are generally circle-like, the larger CirRato is better for decision-making, and the calculation formula is as follows.

$$\text{CirRato} = \frac{4\pi S}{L^2} \tag{13}$$

Where S is the area of the MaskImg, L is the perimeter of the MaskImg contour, and $\text{CirRato} \in [0,1]$.

(2) The boundary adhesion degree is called BAdh, which is the degree of adhesion of the mask image MaskImg to the boundary of image. Since the leukocytes are usually located at the center of the sub graph, they are not connected to the boundary. Therefore, the smaller the BAdh, the better the decision. The calculation formula is as follows.

$$BAdh = (\frac{UpBA + DnBA}{W} + \frac{LtBA + RtBA}{H}) / Acc \tag{14}$$

Where W and H represent the width and height of the mask image, respectively. UpBA, DnBA, BtBA, and RtBA represent the number of pixel points on the upper, lower, left and right boundaries, respectively. Acc indicates how many borders there are pixels, and $BAdh \in [0,1]$.

(3) The grey mean value of saturation channel image S is abbreviated as Sgmv. Pay attention to that the area occupied by the nucleus does not participate in the calculation. Since the grey value of leukocyte is usually greater than the red blood cells and background, the larger Sgmv is more conducive for decision-making. The formula is as follows.

$$Sgmv = \frac{SValue}{NKerA} \tag{15}$$

Where NKerA represents the number of the pixel with the grey value of 255 in image MaskIng, and the grey value of the pixel at the corresponding position is 0 in the image kernelImg, SValue represents the grey value of the corresponding position of the S channel image, and $Sgmv \in [0,1]$.

(4) The quasi-similarity is abbreviated as CirSim, which fits a minimum ellipse to the combined mask image HSGMask (It is defined below) using the least-squares method, and then calculates the similarity between each mask image and the minimum ellipse. The calculation formula is as follows.

$$CirSim = 1 - \frac{outA + inA}{refA} \tag{16}$$

Where outA represents the number of the pixel with the grey value of 255 in image MaskIng, and the grey value of the pixel at the corresponding position is 0 in the image HSGMask. The meaning of inA is the opposite of outA. refA represents the number of non-zero pixels of image HSGMask, and $CirSim \in [0,1]$.

The combined mask image HSGMask is constructed as follows.

$$\text{HSGMask}(i, j) = GMask(i, j) + HMask(i, j) + SMask(i, j) \tag{17}$$

Here, GMask (i, j) represents the grey value at the position of the mask image (i, j) obtained from the G image, the HMask (i, j) and SMask (i, j) have similar meanings.

The calculation formula of the final decision value in this paper is defined as follows.

$$Dec = \frac{1}{3 - CirRato - Sgmv - CirSim + BAdh} \tag{18}$$

It should be noted that, when the three candidate images have leukocytes and over-leukocytes, or leukocytes and under-leukocytes exist simultaneously, then only considering the CirRato and BAdh which not can make a good decision. In this context, over-leukocytes mean that the segmented leukocytes which contain other non-leukocyte components (e.g., red blood cells, background), and under-leukocytes are defined as incomplete leukocytes, such as the nucleus.

In order to solve this problem, we propose two decision parameters, namely Sgmv and CirSim described above. For the coexistence of leukocytes and over-leukocytes, the Sgmv of leukocytes is greater than over-leukocytes, and the CirSim of leukocytes is less than over-leukocytes. Although the smaller CirSim is conducive to decision-making, the Sgmv gap between the two cells is greater. For the coexistence of leukocytes and under-leukocytes, the CirSim of leukocytes is greater than under-leukocytes. At this point, the greater CirSim is conducive to decision-making. Although the Sgmv of leukocytes is less than under-leukocytes, the CirSim gap between the two cells is greater.

In summary，we choose that candidate mask with the largest Dec value as the final segmentation result.

Experimental results of candidate image decision are shown in Table 1. Among them, the second line shows the G-channel image of the ROI image, the various parameters used for decision-making and the corresponding candidate mask image; the third and fourth lines have similar meanings, except that they correspond to the H and S channel images of the ROI image; the last line shows the final segmentation effectiveness of leukocyte.

Table 1. Experimental results of candidate image decision.

| | CirRato | BAdh | Sgmv | CirSim | **Dec** | |
|---|---|---|---|---|---|---|
| 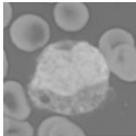 | 0.4986 | 0.0677 | 0.3567 | 0.7121 | **0.6665** | 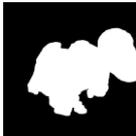 |
| 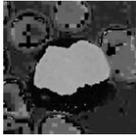 | 0.8020 | 0.0000 | 0.4814 | 0.4607 | **0.7963** | 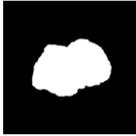 |
| 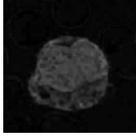 | 0.8561 | 0.0000 | 0.5363 | 0.6994 | **1.1011** | 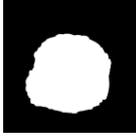 |
| 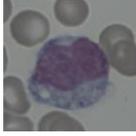 | | choose the largest value of "Dec" → | | | | 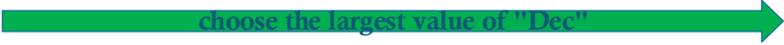 |

# 3. Experiments and results

The details of the experiment are described in more detail below. The experiment is divided into three sections, section 3.1 introduces the experimental environment and sample source, section 3.2 presents the evaluation criteria for the experimental results, and section 3.3 describes the advantages of the proposed method.

## 3.1 Experimental environment

In this paper, the proposed method was implemented using Visual Studio 2010 based on C++ language and experiments conducted using a 64-bit quad-core 3.30GHz Intel i5 processor with a Windows10 operating system and OpenCV2.9.0 visual library.

Two sets of test samples are used in this paper. The first set of test samples comes from Zhongnan Hospital of Wuhan University, about 70853 blood sample images of 100 times magnification, of which 92 had ground truth. The second set of test samples comes from Wuhan Landing Medical High-Tech Co.,Ltd, about 6331 blood sample images of 100 times magnification, of which 11 had ground truth.

Each image in the sample set was taken on an Olympus microscope BX41 (numerical aperture 1.30) with a DFK 23G274 color camera (volume 29 × 29 × 57m3, pixel size 4.4μm × 4.4μm). The difference is that the resolution of the first sample set is 1280×960 pixels; the resolution of the second sample set is 1600×1200 pixels.

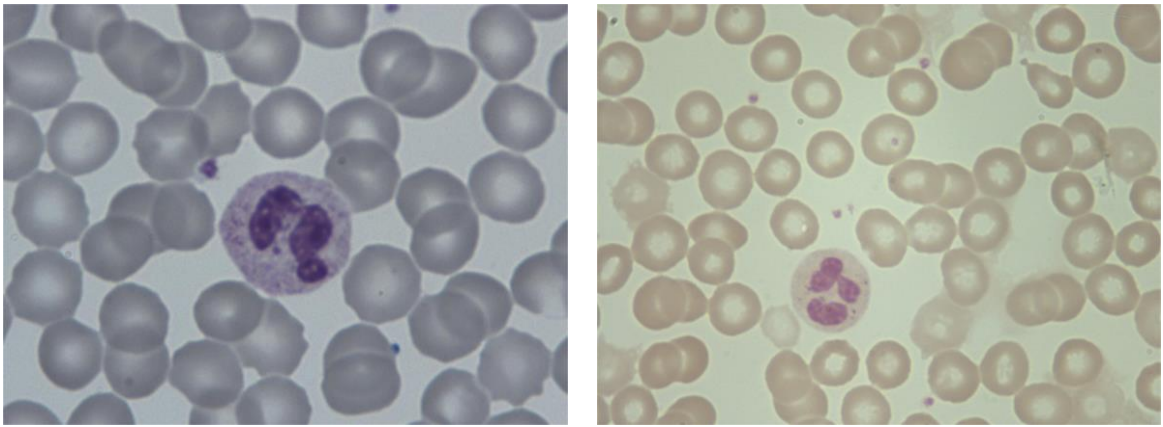

Fig. 6 original microscopic RGB image. The left side is from Zhongnan Hospital of Wuhan University, and the right side is from Wuhan Landing Medical High-Tech Co.,Ltd.

Fig.6 shows the original blood sample image in the RGB format, which was obtained from human peripheral blood. The left side is the blood sample from Zhongnan Hospital of Wuhan University, and the right side is the blood sample from Wuhan Landing Medical High-Tech Co.,Ltd.

## 3.2 Evaluative criteria

To evaluate the segmentation performance of the algorithm, we use the following four different metrics.

(1)Segmentation Accuracy (SA): Fig.7 (f-h) shows the performance of a true-segmented contour superimposed on a segmented image sketched by three specialists. The calculation formula is defined as follows [30].

$$SA = (1 - \frac{|R_s - T_s|}{R_s}) \times 100\% \qquad (19)$$

Where the cell area $R_s$ is that the expert has sketched out manually, the cell area $T_s$ is obtained by segmentation algorithm, and numerator $|R_s - T_s|$ indicates the number of misclassified pixels.

(2) Over-segmentation rate (OR): The formula is defined as follows [43].

$$OR = \frac{O_s}{R_s + O_s} \qquad (20)$$

Where $O_s$ represents the number of pixels that should not be included in the segmented result sets but are included, and $R_s$ has the same meaning as above.

(3) Under-segmentation rate (UR): The formula is defined as follows [43].

$$UR = \frac{U_s}{R_s + O_s} \qquad (21)$$

Where $U_s$ represents the number of pixels that should be included in the segmented result set, but actually not, and $R_s, O_s$ have the same meaning as above.

(4) Error rate (ER): The calculation formula is defined as follows [43].

$$ER = \frac{O_s + U_s}{R_s} \qquad (22)$$

## 3.3 Experimental analysis

Using the algorithm proposed in this paper on Fig. 7(a), the results of cytoplasmic segmentation shown in Fig. 7(e) below. 7(a) shows the original blood smear ROI region, 7(b-d) represents the segmentation results manually drawn by three experts, respectively, 7(e) represents the segmentation results of the algorithm proposed, 7(f-h) shows the performance of superimposing the contours of (e) on (b-d), respectively, the purpose of which is to calculate the segmentation accuracy.

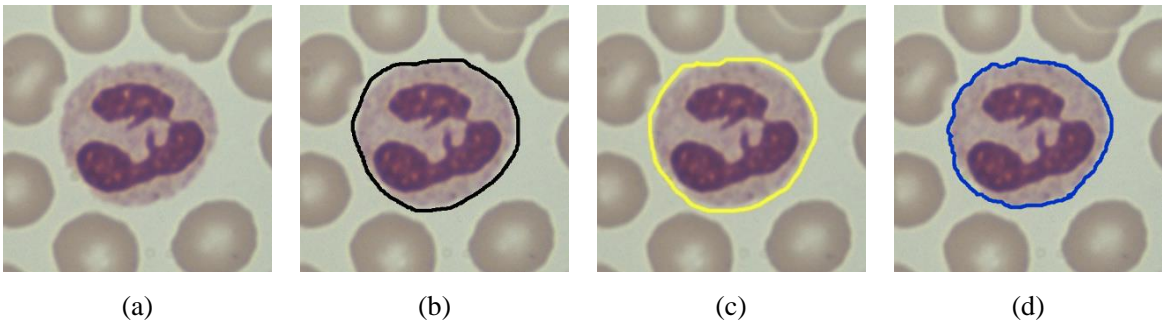

(a)          (b)          (c)          (d)

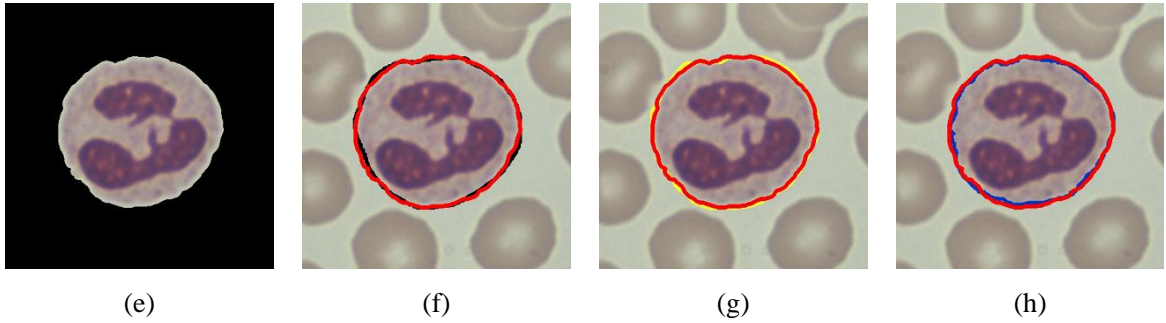

(e) (f) (g) (h)

Fig. 7. (a)The original blood smear image, (b-d) ground truth segmentation results obtained by expert 1, expert 2 and expert 3, respectively, (e) segmentation result of the proposed algorithm, (f-h) the segmented image of (e) is superimposed on the ground truth images to calculate the segmentation accuracy.

Table 2 shows the values of each measured parameters, which was taken by the SWAM &IVFS algorithm in 203 reference sample sets which were drawn by three hematologists.

From the Table 2, we can see the mean segmentation accuracy was 93.75% and the standard deviation (SD) was 0.5984%. The reference sample set were chosen from two sets of test samples mentioned above. The proportion of cells of each type was 10.34% of monocytes, 24.14% of lymphocytes, 60.59% of neutrophils, 3.94% of eosinophils, and 0.99% of basophils.

Table 2. The value of each measured parameters are obtained by applying the algorithm. Note that, "SD" denotes standard deviation.

|  | Ground truth images | | | Mean | SD |
|---|---|---|---|---|---|
|  | Expert-1 | Expert-2 | Expert-3 |  |  |
| Average SA | 93.87% | 92.96% | 94.41% | 93.75% | 0.5984% |
| Average OR | 1.68% | 2.37% | 2.75% | 2.267% | 0.4429% |
| Average UR | 4.33% | 4.49% | 2.66% | 3.827% | 0.8275% |
| Average ER | 6.13% | 7.04% | 5.59% | 6.253% | 0.5984% |

Fig. 8 shows the segmentation results of five types of leukocytes using the SWAM & IVFS algorithm in different environments, where each image uses a red line to depict the profile of the cell, and the neutrophils are subdivided into neutrophilic stab granulocyte and neutrophilic split granulocyte.

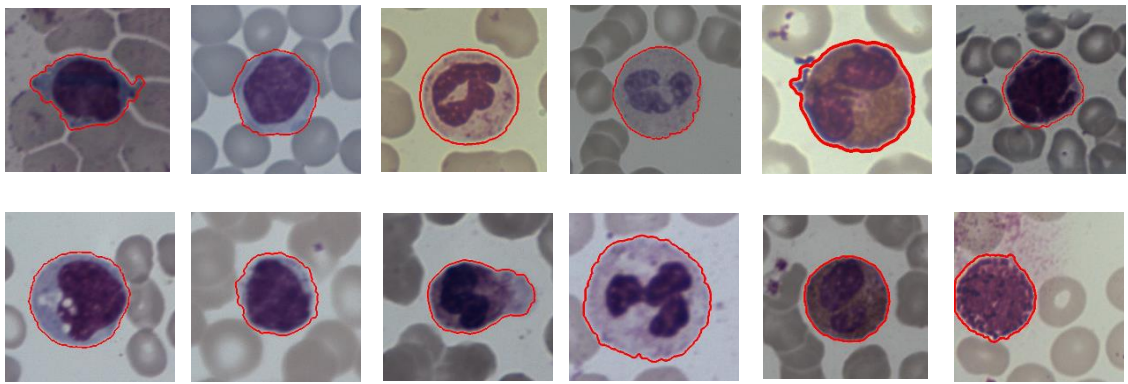

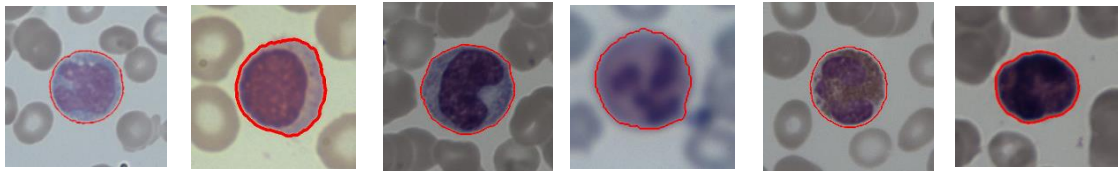

Monocyte    Lymphocyte    NSBG    NSTG    Eosinophil    Basophilic

Fig. 8. The segmentation results of five types of leukocytes in different environments. Note that, "NSBG" denotes neutrophilic stab granulocyte; "NSTG" denotes neutrophilic split granulocyte.

# 4. Discussion

The contrastive experiments in this paper will be divided into two categories, one is comparison with the fuzzy set method, and the other is comparison with the non-fuzzy set method. The experimental results are discussed in section 4.1 and 4.2 respectively.

## 4.1 Fuzzy set method

Fig. 9 shows the comparison between the proposed SWAM&IVFS method and the other three fuzzy set type methods. Among them, the first column shows the reference image which sketched by a medical expert; the second column shows the segmentation results of SWAM&FS method; the third column shows the segmentation results of SWAM&IFS method; the fourth column shows the segmentation results of SWAM&IVIFS method, and the fifth column shows the segmentation results of the proposed method. It should be noted that, SWAM&FS is the method of changing the IVFS in the proposed method to FS. It should be noted that SWAM&FS is a method of changing the IVFS in the proposed method to FS, and SWAM&IFS and SWAM&IVIFS have similar meanings.

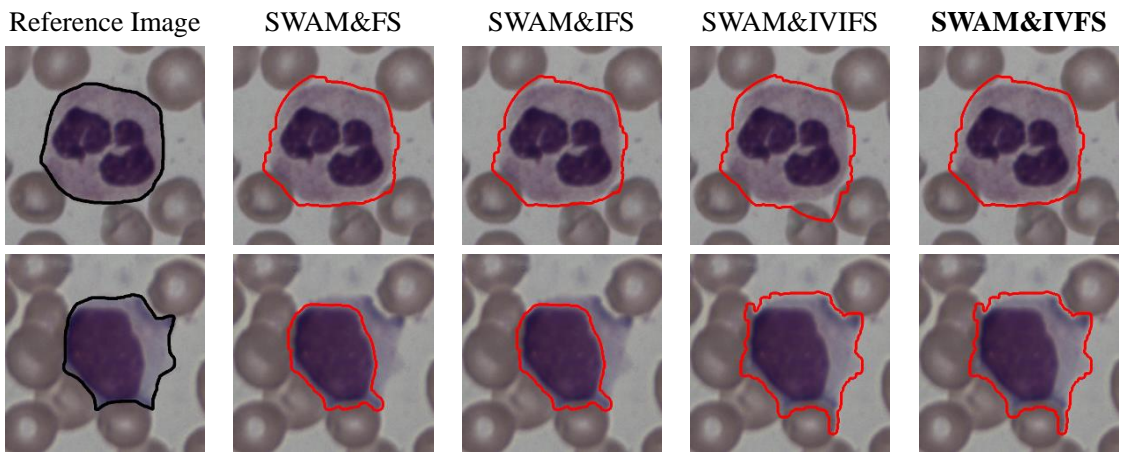

Reference Image    SWAM&FS    SWAM&IFS    SWAM&IVIFS    **SWAM&IVFS**

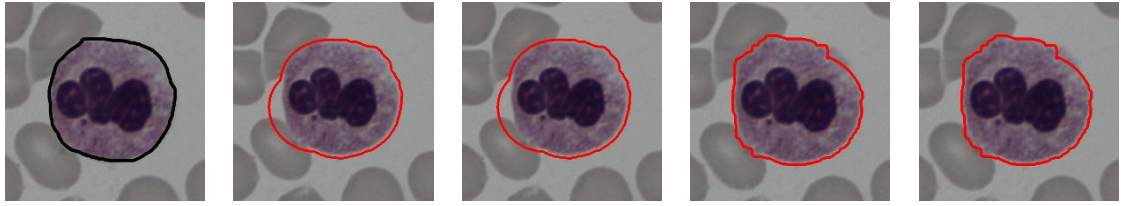

Fig. 9. Contrast with fuzzy sets method. Note that, "FS" denotes the fuzzy sets; "IFS" denotes the intuitionistic fuzzy sets; "IVIFS" denotes the interval-valued intuitionistic fuzzy sets.

Table 3 shows the results of the comparison with the three fuzzy sets methods.

Table 3. The evaluation criteria for the segmentation performance of 203-reference sample

| Methods | Average SA | Average OR | Average UR | Average ER |
|---|---|---|---|---|
| SWAM&FS | 0.9173 | 0.0375 | 0.0388 | 0.0827 |
| SWAM&IFS | 0.9218 | 0.0358 | 0.0375 | 0.0782 |
| SWAM&IVIFS | 0.9225 | 0.0365 | 0.0359 | 0.0775 |
| Ours | 0.9375 | 0.0227 | 0.0383 | 0.0625 |

We know that fuzzy sets, intuitionistic fuzzy sets, interval-valued fuzzy sets, and interval-valued intuitionistic fuzzy sets are further extensions or blurring of classical sets. It can be seen that interval-valued fuzzy sets have better segmentation performances than fuzzy sets, intuitionistic fuzzy sets and interval-valued intuitionistic fuzzy sets from Fig. 9 and Table 3.

## 4.2 Non-fuzzy set method

Table 4 shows the comparison between the proposed SWAM&IVFS method and the other six non-fuzzy set type methods. Among them, "ATPIS" represents the average time required to process an image, in seconds.

Table 4. Comparison on of speed and accuracy on 203-reference sample.

| Methods | Average SA | Average OR | Average UR | Average ER | ATPIS |
|---|---|---|---|---|---|
| Salem [16] | 0.8579 | 0.0122 | 0.1239 | 0.1419 | 1.7395 |
| Liu [17] | 0.8803 | 0.0050 | 0.0978 | 0.1034 | 33.9865 |
| Tareef [27] | 0.8768 | 0.0708 | 0.0405 | 0.1232 | 33.6840 |
| Li [26] | 0.8281 | 0.0068 | 0.1634 | 0.1719 | 1.1652 |
| Song [19] | 0.9025 | 0.0048 | 0.0923 | 0.0975 | 53.0395 |
| Liu [28] | 0.9040 | 0.0046 | 0.0918 | 0.0960 | 6.7300 |
| Ours | 0.9375 | 0.0227 | 0.0383 | 0.0625 | 1.4569 |

In order to see the advantages and disadvantages of each method intuitively, we draw a scatter diagram of accuracy and speed. An excellent segmentation algorithm should be located in the upper left corner of the diagram. As shown in Figure 10, our methods perform well both in speed and accuracy.

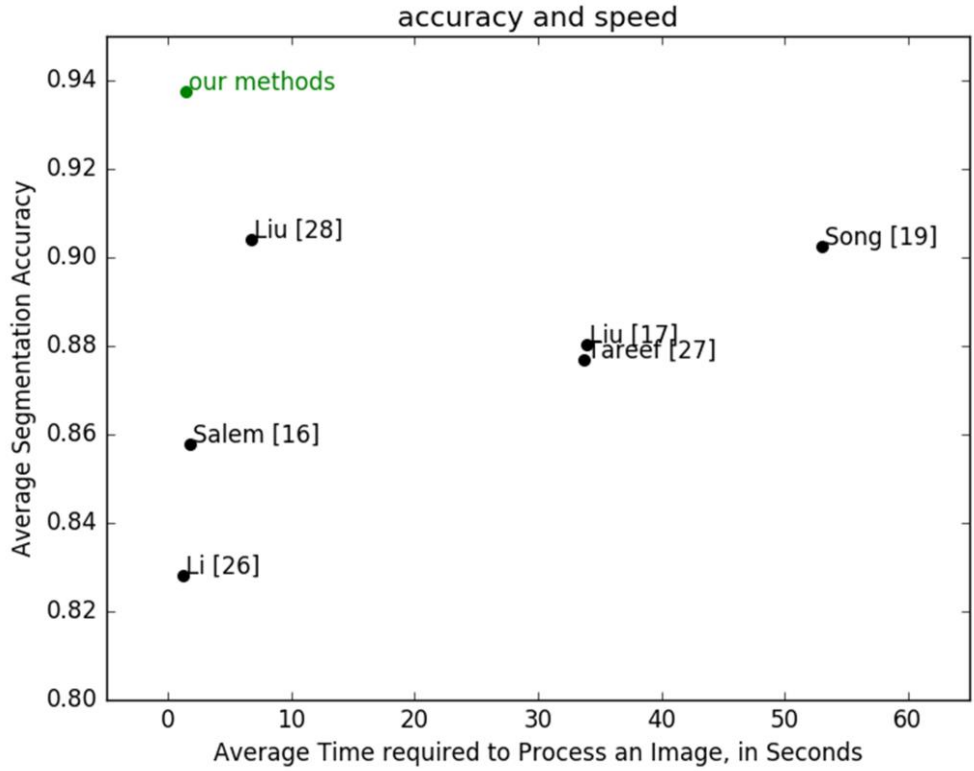

Figure 10. Accuracy and speed on 203-reference sample

# 5. Conclusion

Based on the transformed images, the nucleus was segmented by stepwise averaging method, the cytoplasm was segmented by interval-valued fuzzy sets and post-processing was carried out by using the concave-convex iterative repair algorithm. Meanwhile, in order to adapt to a variety of complex background, we consider the G, H, S three single-channel images. Therefore, a fast and accurate algorithm for peripheral blood leukocyte segmentation was proposed. Experimental results indicate that our proposed method obtains significant improvements over the comparable state-of-the-art segmentation methods.

As previously mentioned, one future work is to improve the situation of cell adhesion. For that, one could consider making use of the canny operator to find the contour of the gradient image obtained by the sobel operator, and then find the missing edge according to the position of the pole on the gradient image. The searching range can be set at the rectangular region determined by the two poles. If the missing contours exist, then the missing contours can be found if the degree of patching is less than a quarter circle. Otherwise, CCIR will be used again to process the missing contours. As another future work, one might consider eliminating false positive cells based on the texture features extracted from the segmented cells.

# Acknowledgements

The National Natural Science Foundation of China (Grant Nos. 61370179 and Grant Nos. 61370181) supported this work. Author would like to acknowledge Zhongnan Hospital of Wuhan University and Wuhan Landing Medical High-Tech Company, limited for providing the test samples. In particular, the author would like to thank the reviewers for their constructive comments in improving the quality of the manuscript.

**Cao Hai Chao** received a bachelor's degree from Nanchang Hangkong University in 2015. Afterwards, he entered Huazhong University of Science and Technology for a Ph.D. degree. His research areas include digital image processing, medical image analysis and deep learning.

**Liu Hong** received Ph.D. in electrical engineering and computer science from Teesside University in the United Kingdom, in 2000. she was hired as a professor and doctoral supervisor in the School of Computer Science at Huazhong University of Science and Technology in 2012. Her research areas include digital image processing, medical image analysis and pattern recognition.

**Song En Min** received a Ph.D. in electrical engineering and computer science from Teesside University in the United Kingdom, in 2000. He was selected as a national "Thousand Talents Program" expert in 2014 and now serves as the director of the digital media at Huazhong University of Science and Technology. His research areas include digital image processing, medical image analysis, artificial intelligence.